\documentclass[11pt]{article}
\usepackage[final]{acl}
\usepackage{times}
\usepackage{latexsym}
\usepackage[T1]{fontenc}
\usepackage[utf8]{inputenc}
\usepackage{microtype}
\usepackage{inconsolata}
\usepackage{graphicx}
\usepackage{amsmath}
\usepackage{booktabs}
\usepackage{xcolor}
\usepackage{colortbl}
\usepackage{afterpage}
\usepackage{enumitem}
\definecolor{resultgroup}{RGB}{235,235,235}
\definecolor{resultours}{RGB}{224,242,248}
\definecolor{resultgain}{RGB}{46,125,50}

\title{RSI-Prime: Training Models to Train Models}
\author{
  Dewu Zheng\textsuperscript{1} \quad
  Yang Ye\textsuperscript{2} \quad
  Hongyu Zhang\textsuperscript{3} \quad
  Jun Yu\textsuperscript{4}  \quad
  Yanlin Wang\textsuperscript{1}\thanks{Yanlin Wang is the corresponding author.} \\
  \textsuperscript{1}Sun Yat-sen University \quad
  \textsuperscript{2}Mind Lab \quad 
  \textsuperscript{3}Chongqing University \\
  \textsuperscript{4}The School of Intelligence Science and Engineering, \\
  Harbin Institute of Technology, Shenzhen \\
  \texttt{zhengdw5@mail2.sysu.edu.cn}, \texttt{yanye@umich.edu} \\
  \texttt{hyzhang@cqu.edu.cn}, \texttt{yujun@hit.edu.cn} \\
  \texttt{wangylin36@mail.sysu.edu.cn}
}

\begin{document}
\maketitle

\begin{abstract}
Enabling large language models (LLMs) to autonomously conduct training and improve target models is an important foundation for recursive self-improvement (RSI). As LLMs expand their role from being trained to conducting training, their model-training capability warrants dedicated study and systematic development. To this end, we construct \textbf{RSI-Prime}, a dataset for improving model-training capability, comprising 350 complete experimental trajectories across seven task categories. To help LLMs learn to make high-quality experimental decisions from training practice, we perform fine-grained decision identification and quality annotation, then apply decision-level loss masking to exclude supervision on low-quality behavior. Experiments with GLM-5.3 on PostTrainBench, a benchmark for autonomous post-training, show that training on RSI-Prime yields a 31.95\% relative improvement in overall performance over the original GLM-5.3, highlighting a path toward future RSI systems in which models autonomously sustain their own advancement.
\end{abstract}

\section{Introduction}
\label{sec:introduction}

\begin{figure*}[!t]
  \centering
  \includegraphics[width=\textwidth]{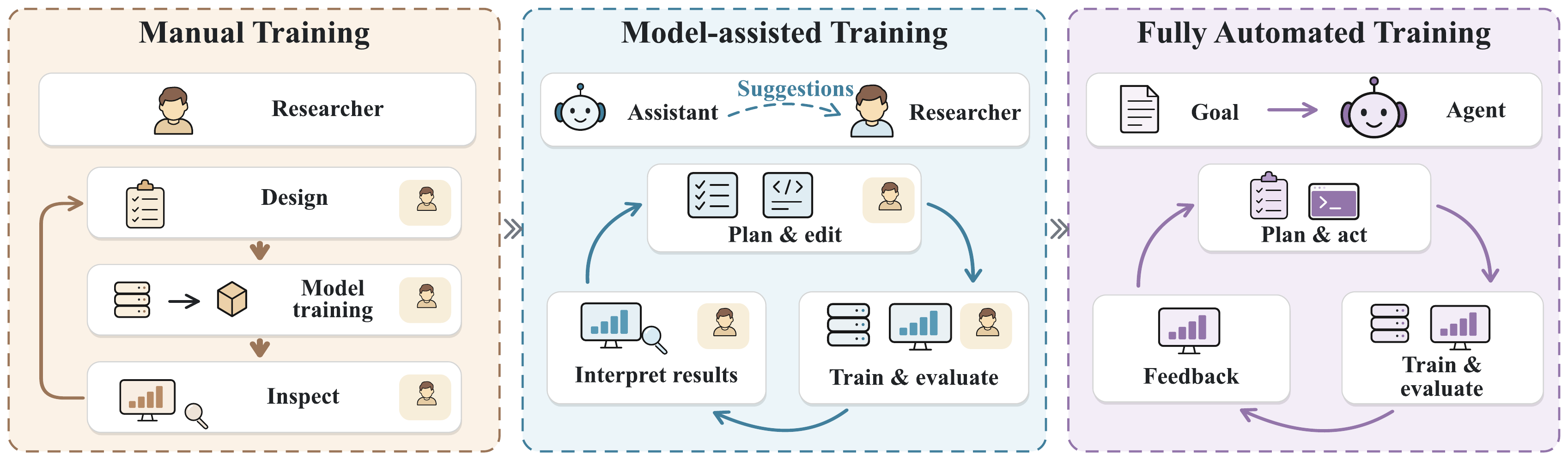}
  \caption{Three paradigms of model training: manual, model-assisted, and fully automated.}
  \label{fig:training-paradigms}
\end{figure*}

Recent large language models (LLMs) have demonstrated growing capabilities in planning and executing model-training experiments~\citep{huang2023mlagentbench,chan2024mlebench,rank2026posttrainbench,yu2026autotrainess}. As illustrated in Figure~\ref{fig:training-paradigms}, model training is expanding from researcher-led manual training to model-assisted training, with the prospect of LLMs autonomously planning, executing, and iterating on experiments toward human-specified goals~\citep{yamada2025aiscientistv2}. As LLMs extend their role from being trained to conducting training, their model-training capability warrants dedicated study and systematic development, laying a foundation for recursive self-improvement (RSI)~\citep{gao2025selfevolving,zhang2026hyperagents}.

Existing research advances model-training automation through improvements to execution harnesses~\citep{trirat2024automlagent,fang2025mlzero,yu2026autotrainess,du2026rsimaster}, data construction~\citep{du2026datamaster,zhao2026andes,zhao2026godelforest}, and training algorithms~\citep{jiang2025aide,ma2026trex}. These advances expand the tools and methods available to agents, but their effective use requires LLMs to select suitable training approaches and adapt them to task requirements and experimental feedback~\citep{lim2026missing,yao2026improveanytask}. This motivates learning from training experience to systematically improve LLMs' own model-training capability across tasks.

To this end, we construct \textbf{RSI-Prime}, a dataset designed to improve model-training capability, comprising 350 complete experimental trajectories across seven task categories. These trajectories preserve the full process of planning, training, evaluation, and iterative refinement, providing context for learning from experimental experience. However, long-horizon training involves many experimental decisions, and final success does not guarantee the quality of each decision. To derive high-quality supervision from complete trajectories, we perform fine-grained identification and quality annotation of experimental decisions, then apply decision-level loss masking based on these annotations. The resulting supervision emphasizes high-quality behavior while preserving the full experimental context, enabling LLMs to learn to make high-quality experimental decisions from training practice.

We conduct experiments with GLM-5.3 on PostTrainBench, a benchmark that assesses an agent's model-training capability through the performance of the target model it trains under bounded resources~\citep{rank2026posttrainbench,posttrainbench2026v12}. Across six tasks, full-trajectory fine-tuning raises the agent's overall score from 36.56 to 45.61, while decision-level supervision further improves it to 48.24, a 31.95\% relative gain over the original GLM-5.3 (Section~\ref{sec:main-results}). Comparisons across supervision granularities highlight the effectiveness of decision-level supervision, which outperforms step-level selection at similar supervision retention levels (Section~\ref{sec:supervision-strategies}). Furthermore, behavioral analysis reveals improvements in decision quality across multiple stages of experimentation, indicating that the gains in task performance are accompanied by higher-quality experimental behavior (Section~\ref{sec:agent-behavior}).

Our main contributions are as follows:
\begin{itemize}[leftmargin=*]
  \item \textbf{A dataset for improving model-training capability.} We construct RSI-Prime, comprising 350 complete experimental trajectories across seven task categories, with environment validation and quality filtering to support learning from training practice.
  \item \textbf{Decision-level supervision for experimental trajectories.} We identify experimental decisions and annotate their action types and quality, enabling selective supervision of high-quality behavior while preserving the full context.
  \item \textbf{Empirical validation on PostTrainBench.} Fine-tuning GLM-5.3 on RSI-Prime yields a 31.95\% relative improvement over the original model. Supervision-strategy comparisons and behavioral analyses further demonstrate the effectiveness of decision-level supervision.
\end{itemize}

\section{Related Work}
\label{sec:related-work}

\subsection{Self-Evolving Agents}
\label{sec:related-self-evolving}

Self-evolving agents use interaction experience and environmental feedback to improve their memory, skills, programs, or model parameters~\citep{gao2025selfevolving}. For experience accumulation, Reflexion converts feedback into reflective memory, ExpeL extracts reusable knowledge from task experience, and ACE continually accumulates and refines contextual strategies~\citep{shinn2023reflexion,zhao2023expel,zhang2025ace}. Voyager and Memento-Skills further support subsequent task execution by building and updating skill libraries, while Agent-Pro improves decision-making through reflection on and revision of beliefs, together with policy search~\citep{wang2023voyager,zhou2026mementoskills,zhang2024agentpro}.

Another line of work extends the object of improvement to agent programs and execution systems. ADAS uses a meta agent to program and search for agent implementations, while the Darwin G\"odel Machine and SICA improve capabilities by modifying their own code and evaluating candidate versions~\citep{hu2024adas,zhang2025dgm,robeyns2025sica}. Self-Harness proposes and validates harness modifications based on execution failures, and The Hyperagents framework further makes the meta-level mechanism that generates improvements editable~\citep{zhang2026selfharness,zhang2026hyperagents}.

Parameter learning uses experience to adapt and train the model itself. SEAL learns to generate fine-tuning data and update directives, while Agent Learning via Early Experience derives supervision from future states produced by the agent's own interactions~\citep{zweiger2025seal,zhang2025earlyexperience}. Self-Challenging trains agents through reinforcement learning on self-generated verifiable tasks, while AgentEvolver combines task generation, experience-guided exploration, and contribution attribution to improve learning efficiency~\citep{zhou2025selfchallenging,zhai2025agentevolver}. RSI-Prime focuses on model-training capability, turning complete experimental processes in which agents train target models into learning resources to improve their ability to conduct subsequent training experiments.

\subsection{Automated LLM Post-Training}
\label{sec:related-post-training}

Automated LLM post-training studies how agents autonomously prepare data, train models, and iterate on evaluation feedback for a target task. PostTrainBench evaluates end-to-end model improvement under bounded resources~\citep{rank2026posttrainbench}, Curation-Bench and RSIBench-Data focus on training-data strategy design and iteration~\citep{kang2026curationbench,meng2026rsibenchdata}, and Agent$^2$ RL-Bench further examines agents' ability to construct and execute reinforcement-learning post-training pipelines, including closed-loop online interaction~\citep{chen2026agent2rlbench}. These benchmarks characterize autonomous model-training capability at different levels.

Existing methods advance autonomous post-training through data, strategy, and execution workflows. DataMaster and ANDES explore data discovery and synthesis~\citep{du2026datamaster,zhao2026andes}, TREX explores training approaches through tree-based experimental search~\citep{ma2026trex}, and G\"odel Forest combines co-evolving search trees with shared experience to balance the depth and breadth of data-strategy exploration~\citep{zhao2026godelforest}. AutoTrainess, RSI-Master, and ImproveAnyTask support experimentation through workflow organization, experiment management, and feedback-driven strategy iteration~\citep{yu2026autotrainess,du2026rsimaster,yao2026improveanytask}. \citet{lim2026missing} further find that the agents they study can repair implementations and repeatedly execute training, yet rarely revise their initial strategies in response to feedback, highlighting the need to improve experimental decision-making.

Learning from training experience offers another path to improving agents' autonomous post-training capability. RSI-Dojo constructs post-training environments, collects and filters experimental trajectories, and uses them to fine-tune post-training agents~\citep{gao2026rsidojo}. RSI-Prime advances this direction through a systematic pipeline for reliable trajectory construction and fine-grained supervision in long-horizon experiments. Pre-execution validation, independent experimental supervision, and interruption recovery that preserves the original session and cumulative budget support reliable, traceable trajectory collection. Fine-grained decision identification and quality annotation then guide decision-level loss masking, enabling selective supervision of high-quality behavior while preserving the full experimental context.

\begin{figure*}[!t]
  \centering
  \includegraphics[width=\textwidth]{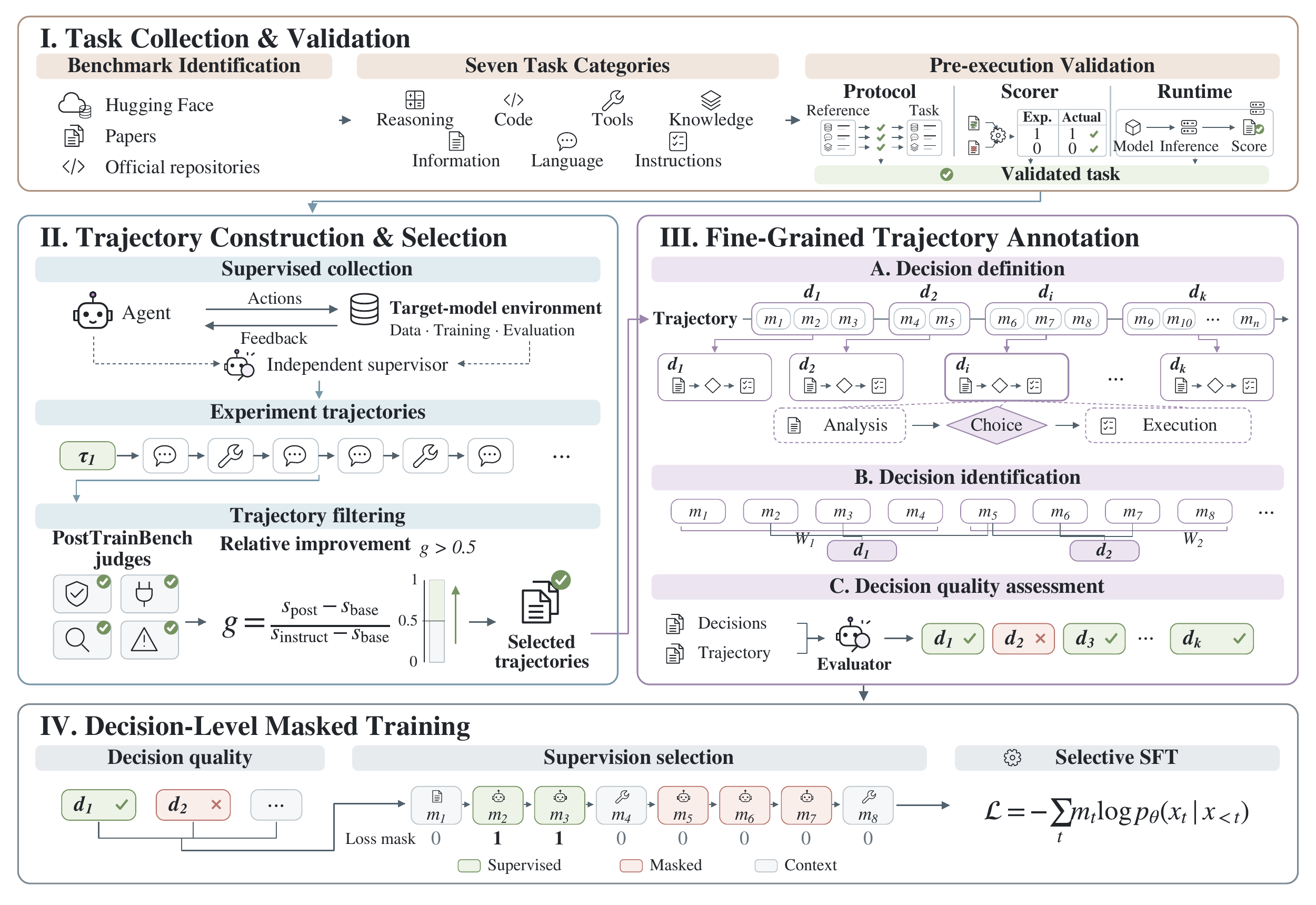}
  \caption{Overview of the RSI-Prime pipeline.}
  \label{fig:data-construction}
\end{figure*}

\section{Methodology}
\label{sec:methodology}

\subsection{Overview}
\label{sec:overview}

RSI-Prime provides supervision for improving agents' model-training capability by constructing complete trajectories of model-training experiments and selecting high-quality behavior within them. As shown in Figure~\ref{fig:data-construction}, the pipeline comprises four stages: task collection and validation, trajectory construction and selection, fine-grained trajectory annotation, and decision-level masked training.

We first collect benchmarks across seven task categories, integrate them into a common interface, and validate their evaluation protocols, scorers, and runtime workflows to establish reliable task environments for trajectory collection. In these environments, agents conduct training experiments under independent supervision. The resulting trajectories are filtered for protocol compliance and target-model improvement before proceeding to fine-grained annotation. We identify experimental decisions within these trajectories and annotate their quality and action types using source evidence and context, supporting supervision selection and behavioral analysis, respectively. Quality judgments are then mapped to message-level loss masks for selective supervised fine-tuning while preserving the full experimental context.

\subsection{Task Collection and Validation}
\label{sec:task-collection-validation}

We collect and integrate benchmarks into a common task interface, then validate their evaluation protocols, scorers, and runtime environments before trajectory collection.

\paragraph{Benchmark identification.}
\label{sec:benchmark-identification}
To cover diverse model-training objectives, we automatically collect candidate benchmarks from Hugging Face, academic papers, and official repositories across seven task categories: mathematics and reasoning, code, tools and interaction, knowledge and understanding, information processing, language and text, and instruction following~\citep{liang2023helm,wang2022supernaturalinstructions}. The pipeline consults original papers, dataset documentation, and implementations to extract task definitions, data splits, and evaluation protocols, then integrates selected benchmarks into a common task interface. This interface specifies the target model, resource budget, and submission requirements, standardizing execution and result recording while preserving each benchmark's scoring semantics and providing task environments for pre-execution validation.

\paragraph{Pre-execution validation.}
\label{sec:pre-execution-validation}
To ensure reliable experimental feedback, we perform three types of validation shown in Figure~\ref{fig:data-construction}. \textbf{Protocol validation} checks task scope, model identity, prompts, generation settings, and the primary metric against the original evaluation requirements. \textbf{Scorer validation} replays available Base/Instruct outputs and uses positive and negative controls and coverage checks to verify expected scoring behavior. \textbf{Runtime validation} tests the supplied model's inference and scoring pipeline on the actual execution host. Tasks must pass all checks before trajectory collection; failures require repair and revalidation.

\subsection{Trajectory Construction and Selection}
\label{sec:trajectory-construction}

We collect complete model-training trajectories through independent supervision and interruption recovery, then filter them for protocol compliance and target-model improvement. The trajectories retain successful and unsuccessful intermediate attempts, providing context for decision identification and quality assessment.

\paragraph{Supervised collection.}
\label{sec:supervised-experiments}
Maintaining reliable execution is challenging in long-horizon training experiments, where agents may encounter interruptions or terminate prematurely~\citep{rank2026posttrainbench,du2026rsimaster}. We use an independent supervisor to monitor experimental progress and a controller to resume interrupted or prematurely terminated runs within the remaining budget. Recovery preserves the original session, artifacts, and cumulative execution time, while supervisor interventions are recorded separately from agent actions. Together, these mechanisms support stable and continuous operation of the trajectory-collection framework.

\paragraph{Trajectory filtering.}
We filter trajectories for protocol compliance and training effectiveness, retaining experimental experience that meets task requirements and substantially improves the target model. Each trajectory must pass PostTrainBench's four trajectory judges, and its submitted model must satisfy the relative-improvement threshold:
\begin{equation}
  g = \frac{s_{\mathrm{post}}-s_{\mathrm{base}}}
           {s_{\mathrm{instruct}}-s_{\mathrm{base}}} > 0.5,
  \label{eq:trajectory-selection}
\end{equation}
where $s_{\mathrm{post}}$, $s_{\mathrm{base}}$, and $s_{\mathrm{instruct}}$ are the scores of the submitted, base, and reference instruction-tuned models under the same benchmark protocol, respectively. This condition requires the submitted model's improvement to exceed half the reference Instruct--Base performance gap.

\subsection{Decision Identification}
\label{sec:decision-identification}

Each experimental trajectory comprises a sequence of decisions that guide the training process from initial planning through iterative refinement. We define an experimental decision as a choice that determines the course of action in a model-training experiment, such as selecting training data, designing a training approach, or revising a strategy based on evaluation feedback. As shown in Figure~\ref{fig:data-construction}, analysis explains the basis for a choice, the choice specifies the action to take next, and execution records how that choice is implemented and the resulting feedback. Together, Analysis, Choice, and Execution capture the progression from rationale to action. We organize content concerning the same choice into one decision, allowing a decision to span multiple messages and a single message to involve multiple decisions.

\paragraph{Decision identification.}
We use an annotation agent to identify decisions in experimental trajectories and link them to their supporting analysis and execution. The agent reads the trajectory in windows that preserve complete interaction rounds, retrieves original messages to recover relevant context, and links related content across windows to the same decision. Content whose association remains unclear at a window boundary is retained as pending and reconsidered with new evidence in subsequent windows, avoiding forced decision splits or attachment to the preceding decision solely because of a window cutoff.

\paragraph{Iterative verification.}
We check the consistency of decision annotations with the original trajectories through iterative independent verification. Each round comprises three independent checks of the same annotation version, assessing decision completeness, evidential support, and the appropriateness of decision boundaries and evidence attribution. Annotations are retained when at least two checks approve them; otherwise, decisions in the corresponding source interval are re-identified and undergo another verification round until majority approval is reached. Annotation reliability is additionally assessed through an independent human audit (Appendix~\ref{sec:human-validation}).

\subsection{Decision Quality Assessment and Action Annotation}
\label{sec:decision-annotation}

We annotate experimental decisions along two axes, quality and action type, to support supervision selection and behavioral analysis, respectively. Both annotations draw on the original trajectories and experimental context, rather than judging local behavior solely by final scores.

\paragraph{Quality assessment.}
We assess decision quality along three dimensions: Analysis, Choice, and Execution. Analysis concerns whether experimental information is correctly interpreted and provides sufficient grounds for the choice; Choice concerns whether the selected action fits the goals and constraints at the time; Execution concerns whether the choice is faithfully implemented and exceptions are appropriately handled. Assessment distinguishes avoidable behavioral errors from external environmental failures: eventual experimental success does not automatically establish the quality of local decisions, nor does a failed tool call alone establish problematic execution.

We obtain three quality assessments from GLM-5.3 for each decision, aggregate judgments by majority voting within each dimension, and use the aggregated judgments to determine decision quality. As shown in Figure~\ref{fig:data-construction}, assessment outcomes are presented as high-quality or problematic. The binary summary describes decision quality, while training eligibility is determined from the underlying dimension-level judgments using the selection rule in Appendix~\ref{sec:supervision-details}.

\paragraph{Action annotation.}
We combine manual taxonomy construction with model-assisted annotation to characterize the actions involved in experimental decisions. Researchers inspect an initial set of trajectories and use card sorting to derive action categories, resolving ambiguous assignments through discussion and extending the taxonomy to cover previously unrepresented behaviors. The resulting taxonomy comprises five main categories---data preparation, model training, evaluation and diagnosis, model composition and delivery, and runtime support---with 20 specific subcategories and an additional Other category.

Using this taxonomy and the original evidence, the model produces three action annotations for each decision, with the category determined by majority voting. Classification follows the core choice; compound decisions receive the most representative category, while records without sufficient support for classification are assigned to Other. Action labels support dataset analysis and subsequent behavioral comparisons.

\subsection{Decision-Level Masked Training}
\label{sec:decision-masked-training}

We use decision quality to select training supervision, reducing imitation of low-quality behavior while preserving the full experimental context. As shown in Figure~\ref{fig:data-construction}, quality assessments guide message selection, followed by selective fine-tuning through loss masking. Selection is based on decisions, while the mask operates on whole assistant messages, allowing cross-message decision evidence to determine the training content.

\paragraph{Supervision selection.}
We use source references in decision annotations to map analysis, choice, and execution content back to the corresponding assistant messages. Evidence in tool results is linked to the assistant message that initiated the call. Because decisions and messages have a many-to-many relationship, a message may contain content associated with both passing and failing decisions. We adopt a conservative rule: a message is supervised only when all its linked decisions pass quality selection; otherwise, the entire message is masked. Messages without linked decisions are also excluded from supervision.

Let $\mathcal{D}(a)$ denote the decisions linked to assistant message $a$, and define $q(d)=1$ if decision $d$ passes quality selection and $q(d)=0$ otherwise. The message selection mask is
\begin{equation}
  z_a =
  \begin{cases}
    \displaystyle\prod_{d\in\mathcal{D}(a)}q(d), & \mathcal{D}(a)\ne\emptyset,\\
    0, & \mathcal{D}(a)=\emptyset.
  \end{cases}
  \label{eq:decision-message-mask}
\end{equation}
This rule excludes supervision on content associated with failing decisions, but may also mask high-quality content within the same message. The resulting selection retains 76.25\% of originally supervised assistant messages and 75.48\% of supervised tokens.

\paragraph{Selective SFT.}
The message selection mask and the original supervision mask jointly determine which tokens contribute to training, keeping originally unsupervised content at zero loss. Let $b_t$ be the original supervision mask for token $x_t$ and $a(t)$ its assistant message. We set $m_t=b_tz_{a(t)}$, yielding the training objective
\begin{equation}
  \mathcal{L}(\theta)=-\sum_{\tau\in\mathcal{T}}\sum_t
  m_t\log p_\theta(x_t\mid x_{<t}),
  \label{eq:masked-sft}
\end{equation}
where $\mathcal{T}$ contains the training trajectory segments and $\theta$ denotes the agent-model parameters. Loss masking changes only the supervision scope, preserving all original messages and tool interactions. The model can therefore use observations, failures, and corrections to interpret subsequent choices and learn selected behavior within the full experimental context.

\begin{figure*}[!t]
  \centering
  \includegraphics[width=\textwidth]{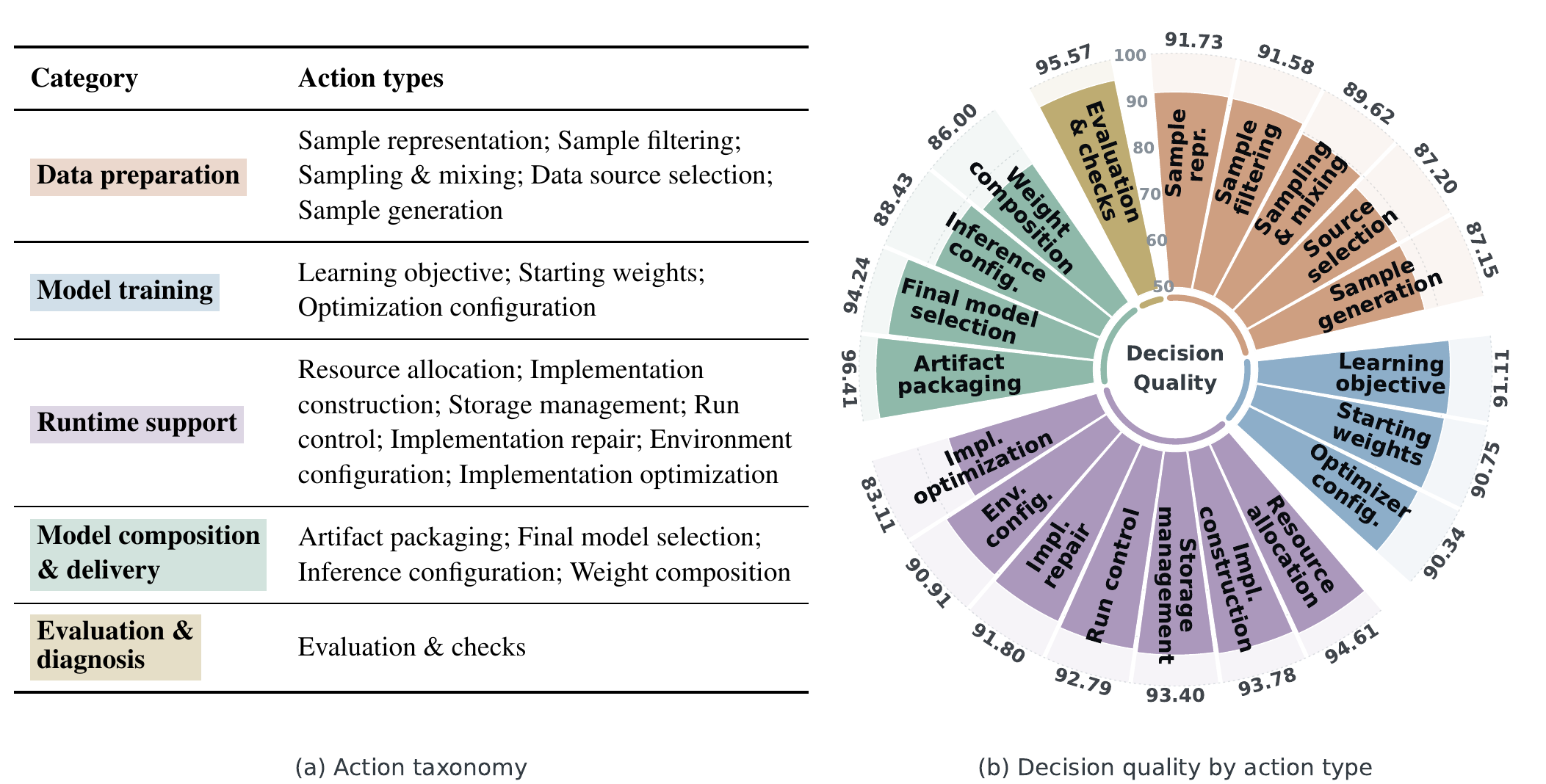}
  \caption{Action taxonomy and decision quality in RSI-Prime.}
  \label{fig:action-quality}
\end{figure*}

\subsection{Dataset Analysis}
\label{sec:dataset-analysis}

We analyze RSI-Prime in terms of experimental activities and decision quality, characterizing the main activities in model training, their quality differences, and the need for supervision selection within complete trajectories. The action taxonomy provides a common perspective for these analyses and subsequent comparisons of agent behavior.

\paragraph{Dataset composition.}
RSI-Prime contains 350 complete experimental trajectories across seven task categories, yielding 385 training segments. The available annotation analysis contains 90,738 decision records, covering five main categories---data preparation, model training, evaluation and diagnosis, model composition and delivery, and runtime support---with 20 specific subcategories and an additional Other category. Evaluation and diagnosis account for 38.43\% of records and data preparation for 27.09\%, together constituting most records. This distribution reflects the iterative nature of training experiments around data construction and evaluation feedback, with experience extending beyond training execution to strategy revision and experimental support.

Decision records and source messages have a many-to-many relationship: 54.95\% of records reference multiple assistant messages, while 57.01\% of referenced assistant messages involve multiple decisions. The analysis, expression, and execution feedback associated with a choice may be distributed across messages, and a single message may discuss several choices.

\paragraph{Decision quality.}
High-quality and problematic decision records account for 92.42\% and 7.58\%, respectively. Most records meet the quality criterion, but quality varies across action types. As shown in Figure~\ref{fig:action-quality}, the proportions of problematic records for implementation optimization and data source selection are 16.89\% and 12.80\%, respectively. Thus, even after trajectory filtering, individual experimental activities differ in supervision quality, and the final outcome alone does not establish the quality of all constituent behaviors. The annotations supporting these analyses were assessed through a human audit, whose procedure and results are provided in Appendix~\ref{sec:human-validation}.

These findings support supervision selection within complete experiments: final performance gains can coexist with local behavior unsuitable for imitation. Decision-level annotation enables such behavior to be identified while preserving the surrounding experimental context, providing a basis for quality-guided selective supervision.

\begin{table}[t]
  \centering
  \small
  \setlength{\tabcolsep}{4pt}
  \begin{tabular}{@{}lrr@{}}
    \toprule
    \textbf{Supervision} & \textbf{Full SFT} & \textbf{Decision SFT} \\
    \midrule
    Messages & 89,211 & 68,027 \\
    Tokens (M) & 44.35 & 33.48 \\
    Message retention (\%) & 100.00 & 76.25 \\
    Token retention (\%) & 100.00 & 75.48 \\
    \bottomrule
  \end{tabular}
  \caption{Supervision before and after decision-level masking.}
  \label{tab:supervision-volume}
\end{table}

\section{Experimental Setup}
\label{sec:experimental-setup}

\begin{table}[!htbp]
  \centering
  \small
  \renewcommand{\arraystretch}{1.08}
  \begin{tabular*}{\columnwidth}{@{\extracolsep{\fill}}lr@{}}
    \toprule
    \textbf{Benchmark} & \textbf{Weight (\%)} \\
    \midrule
    AIME 2025 & 24.48 \\
    Arena Hard Writing & 9.76 \\
    GPQA Main & 24.27 \\
    GSM8K & 10.11 \\
    HealthBench & 19.90 \\
    HumanEval & 11.47 \\
    \bottomrule
  \end{tabular*}
  \caption{PostTrainBench v1.2 benchmark weights, rounded to two decimal places.}
  \label{tab:benchmark-weights}
\end{table}

\begin{table*}[!t]
  \centering
  \small
  \setlength{\tabcolsep}{5pt}
  \renewcommand{\arraystretch}{1.26}
  \resizebox{\textwidth}{!}{%
  \begin{tabular}{ll!{\color{black!25}\vrule width 0.4pt}cccccc!{\color{black!25}\vrule width 0.4pt}l}
    \toprule
    & & \multicolumn{7}{c}{\textbf{PostTrainBench v1.2}} \\
    \cmidrule(lr){3-9}
    \textbf{Model} & \textbf{Agent Harness} & AIME 2025 & Arena Hard Writing & GPQA Main & GSM8K & HealthBench & HumanEval & \textbf{Overall} \\
    \midrule
    \rowcolor{resultgroup}
    \multicolumn{9}{c}{\textbf{Target-model baselines}} \\
    Qwen3-4B-Base & --- & 3.33 & 3.42 & 13.39 & 41.85 & 13.38 & 36.59 & 15.49 \\
    Qwen3-4B (Instruct) & --- & 53.33 & 86.84 & 44.64 & 93.78 & 52.72 & 77.44 & 61.23 \\
    \addlinespace[3pt]
    \rowcolor{resultgroup}
    \multicolumn{9}{c}{\textbf{Closed-source LLMs}} \\
    Fable 5.1 & Claude Code & 25.33 & 93.73 & 37.46 & 90.98 & 53.91 & 81.10 & 53.68 \\
    GPT-6 (Astra) & Codex CLI & 21.00 & 89.34 & 38.43 & 92.49 & 43.09 & 84.45 & 50.81 \\
    Opus 5 & Locus & 20.74 & 86.78 & 37.40 & 90.23 & 45.30 & 82.52 & 50.23 \\
    Gemini 4 Argon & OpenCode & 16.33 & 70.96 & 36.01 & 91.66 & 45.64 & 86.28 & 47.92 \\
    Opus 5.5 & Claude Code & 21.00 & 85.18 & 26.49 & 67.85 & 57.32 & 84.45 & 47.84 \\
    GPT-6.1 (Sol) & Codex CLI & 21.00 & 75.87 & 37.28 & 91.89 & 43.90 & 59.45 & 46.45 \\
    Fable 5 & Claude Code & 21.00 & 85.70 & 30.69 & 90.11 & 35.03 & 80.79 & 46.31 \\
    Opus 5 & Claude Code & 20.33 & 42.48 & 37.83 & 91.66 & 40.95 & 36.59 & 39.92 \\
    \addlinespace[3pt]
    \rowcolor{resultgroup}
    \multicolumn{9}{c}{\textbf{Open-source LLMs}} \\
    GLM-5.2 & Claude Code & 15.56 & 54.25 & 34.10 & 85.77 & 23.86 & 74.66 & 39.37 \\
    Kimi K3 & Claude Code & 20.22 & 38.90 & 34.25 & 86.83 & 20.11 & 79.88 & 39.01 \\
    GLM-5.3 & DSH & 14.44 & 38.46 & 28.94 & 83.65 & 23.58 & 79.27 & 36.56 \\
    \midrule
    RSI-Prime (Full SFT) & DSH & 25.33 & 43.05 & 32.14 & 87.72 & 47.77 & 78.66 & 45.61\textsuperscript{\textcolor{resultgain}{\scriptsize +24.75\%}} \\
    \rowcolor{resultours}
    \textbf{RSI-Prime (Decision SFT)} & DSH & 26.67 & 49.50 & 33.93 & 89.69 & 50.21 & 83.54 & \textbf{48.24}\textsuperscript{\textcolor{resultgain}{\scriptsize +31.95\%}} \\
    \bottomrule
  \end{tabular}}
  \caption{PostTrainBench v1.2 results for the Qwen3-4B target model. Superscripts report relative gains in Overall over the original GLM-5.3 agent.}
  \label{tab:experimental-results}
\end{table*}

\paragraph{Benchmark.}
We use PostTrainBench v1.2 to assess agents' autonomous post-training capability through the performance of the target models they train under bounded resources~\citep{rank2026posttrainbench,posttrainbench2026v12}. Its six tasks cover mathematical reasoning, scientific question answering, writing, health-related responses, and code generation: AIME 2025 and GSM8K assess competition-level and grade-school mathematical reasoning, respectively; GPQA Main assesses graduate-level scientific reasoning; Arena Hard Writing evaluates writing quality; HealthBench evaluates health-related responses; and HumanEval tests the functional correctness of generated code through execution.

\paragraph{Model and training.}
We perform supervised fine-tuning of GLM-5.3 on RSI-Prime for two epochs, with a batch size of 32, a learning rate of $10^{-4}$, a cosine learning-rate schedule, and a warmup ratio of 0.05.

\paragraph{Execution and scoring.}
In each experiment, the target model is initialized from Qwen3-4B-Base. The agent trains and iteratively improves it using one NVIDIA H100 GPU and a cumulative 10-hour execution budget, then submits the final model for scoring; final evaluation is timed separately. The agent uses DeepSeek Harness (DSH) with a 400,000-token context window, a 128,000-token output limit, and maximum reasoning effort. We choose DSH because it supports inspection and modification of its execution mechanisms, facilitating experimental adaptation and maintenance.

We report task scores on a 0--100 scale and the official weighted aggregate:
\begin{equation}
  S=\sum_{b=1}^{6}w_bs_b,
\end{equation}
where $s_b$ is the score on benchmark $b$ and $w_b$ is its official v1.2 weight. The weights are proportional to the inverse of the reference Instruct--Base performance gap and normalized to sum to one~\citep{posttrainbench2026v12}. Aggregation uses full-precision weights; Table~\ref{tab:benchmark-weights} reports their percentages. The training corpus contains HumanEval-related content, resulting in training--evaluation content overlap for this task.

\section{Experimental Results}
\label{sec:experimental-results}

\subsection{Main Results}
\label{sec:main-results}

\textbf{Learning from experimental trajectories improves autonomous post-training capability.} As shown in Table~\ref{tab:experimental-results}, full-trajectory fine-tuning raises the GLM-5.3 agent's Overall score from 36.56 to 45.61, while decision-level supervision further improves it to 48.24, corresponding to relative gains of 24.75\% and 31.95\% over the original GLM-5.3. Full-trajectory supervision already yields substantial gains, and selecting supervision based on decision quality achieves stronger task performance, demonstrating the training value of experimental experience and its fine-grained annotations.

\begin{table*}[!t]
  \centering
  \small
  \setlength{\tabcolsep}{5pt}
  \renewcommand{\arraystretch}{1.12}
  \resizebox{\textwidth}{!}{%
  \begin{tabular}{l!{\color{black!25}\vrule width 0.4pt}cc!{\color{black!25}\vrule width 0.4pt}cccccc!{\color{black!25}\vrule width 0.4pt}l}
    \toprule
    & \multicolumn{2}{c}{\textbf{Supervision retained (\%)}} & \multicolumn{7}{c}{\textbf{PostTrainBench v1.2}} \\
    \cmidrule(lr){2-3}\cmidrule(lr){4-10}
    \textbf{Supervision strategy} & Assistant messages & Tokens & AIME 2025 & Arena Hard Writing & GPQA Main & GSM8K & HealthBench & HumanEval & \textbf{Overall} \\
    \midrule
    Full SFT & 100.00 & 100.00 & 25.33 & 43.05 & 32.14 & 87.72 & 47.77 & 78.66 & 45.61 \\
    Tool-failure masking & 87.69 & 89.15 & 23.33 & 47.73 & 33.93 & 84.84 & 46.17 & 76.22 & 45.12 \\
    Step-level masking & 77.38 & 76.63 & 20.67 & 41.93 & 31.72 & 88.86 & 39.11 & 83.54 & 43.21 \\
    \rowcolor{resultours}
    \textbf{Decision SFT} & \textbf{76.25} & \textbf{75.48} & \textbf{26.67} & \textbf{49.50} & \textbf{33.93} & \textbf{89.69} & \textbf{50.21} & \textbf{83.54} & \textbf{48.24} \\
    \bottomrule
  \end{tabular}}
  \caption{Comparison of supervision strategies on RSI-Prime.}
  \label{tab:supervision-strategies}
\end{table*}

\textbf{Performance gains span diverse training objectives.} Decision SFT improves over the original GLM-5.3 agent on all six tasks, including gains of 12.23 points on AIME 2025 and 26.63 points on HealthBench, alongside improvements in scientific question answering, writing, and code generation. These tasks require agents to organize data, training, and evaluation around different objectives. The cross-task gains indicate that RSI-Prime provides training value across multiple post-training settings.

\textbf{Decision-level supervision further improves full-trajectory fine-tuning.} Decision SFT outperforms Full SFT on all six tasks, increasing Overall by 2.63 points. The largest gains occur on Arena Hard Writing and HumanEval, at 6.45 and 4.88 points, respectively, spanning open-ended writing evaluation and execution-based code testing. These results support using decision quality to select supervision while preserving the full experimental context.

\textbf{RSI-Prime achieves competitive performance among published agents.} Decision SFT reaches 48.24, surpassing Gemini 4 Argon, Opus 5.5, GPT-6.1 (Sol), and Fable 5, and scoring 5.44 points below the SOTA, Fable 5.1 with Claude Code~\citep{posttrainbench2026leaderboard}.

\subsection{Comparison of Supervision Strategies}
\label{sec:supervision-strategies}

To evaluate the effectiveness of selecting supervision at the decision level, we compare Decision SFT with Full SFT and other fine-grained supervision methods, including step-level masking and tool-failure masking. All methods preserve the full training context and select supervision by adjusting loss masks on assistant messages.

\paragraph{Comparison strategies.}
We construct two baselines:
\begin{itemize}
  \item \textbf{Tool-failure masking.} Using explicit execution status, this strategy masks assistant messages associated with tool errors, nonzero exit codes, timeouts, or signal termination. If a message contains multiple tool calls, a failure in any call masks the entire message.
  \item \textbf{Step-level masking.} We adapt the critical-step selection method of ATLAS~\citep{chen-etal-2025-atlas}, which selects supervision according to a step's importance for task completion. We adjust only the step-selection budget, preserving the original importance criterion, to obtain supervision retention close to Decision SFT and compare different granularities and selection criteria at similar supervision volumes.
\end{itemize}

\begin{figure*}[!t]
  \centering
  \includegraphics[width=\textwidth]{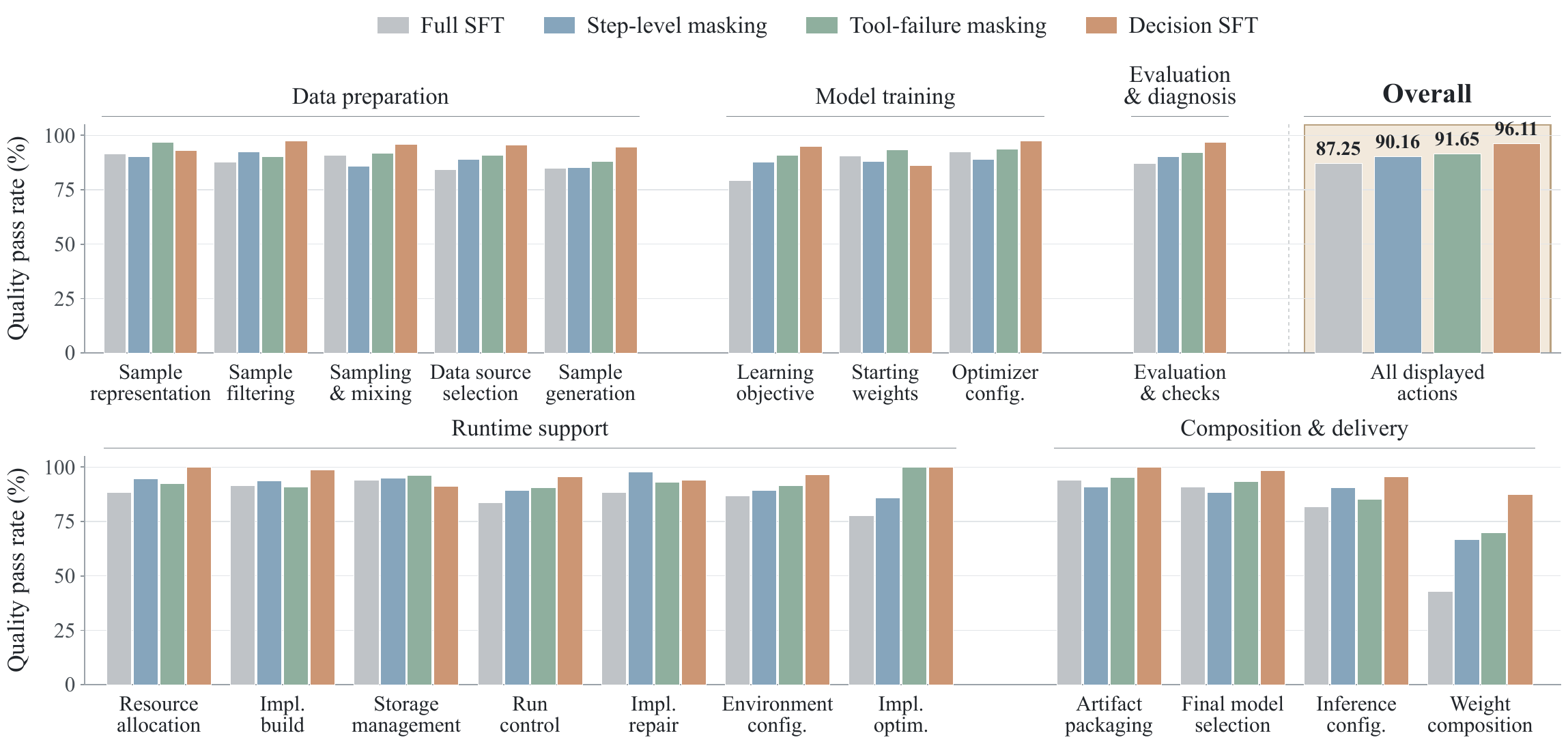}
  \caption{Decision quality by action type across supervision strategies. Bars report quality pass rates for the 20 secondary action types grouped into the five categories used in Figure~\ref{fig:action-quality}. Overall is the fraction of qualified records pooled across these action types; Other is omitted.}
  \label{fig:agent-behavior}
\end{figure*}

\textbf{Decision-level supervision improves performance with less supervision.} As shown in Table~\ref{tab:supervision-strategies}, Decision SFT retains 75.48\% of supervised tokens and achieves an Overall score of 48.24, exceeding Full SFT by 2.63 points and improving all six tasks. Tool-failure and step-level masking score 45.12 and 43.21, respectively, both below Full SFT at 45.61. Reducing supervision alone therefore does not guarantee gains; the granularity and criterion of supervision selection also matter.

\textbf{Decision-level supervision yields more consistent gains at similar supervision volumes.} Step-level masking and Decision SFT retain 76.63\% and 75.48\% of supervised tokens, respectively, yet Decision SFT scores 5.03 points higher overall, outperforming step-level masking on five tasks and matching it on HumanEval. The gains on HealthBench, Arena Hard Writing, and AIME 2025 are 11.10, 7.57, and 6.00 points, respectively. Step-level masking exceeds Full SFT on HumanEval and GSM8K, but these gains do not extend to the other tasks; Decision SFT improves over full supervision on all six. Decision-level annotations connect analysis, choices, and execution distributed across messages, providing more complete experimental evidence for selecting behavior suitable for supervision.

\textbf{Explicit tool-failure signals do not fully capture experimental decision quality.} Tool-failure masking retains 89.15\% of supervised tokens, while Decision SFT uses less supervision yet scores 3.12 points higher overall, outperforming it on five tasks and matching it on GPQA Main. Tool-failure masking improves Arena Hard Writing and GPQA Main over Full SFT, indicating that execution status has some value for selection. However, successful execution does not establish that the selected data, training approach, or evaluation is appropriate for the task. Decision-level assessment additionally considers the supporting analysis and execution feedback to evaluate choices, providing a basis for supervision selection in long-horizon experiments.

\subsection{Agent Behavior Analysis}
\label{sec:agent-behavior}

Beyond final task performance, we further examine whether agents' experimental decision quality also improves. We analyze evaluation trajectories under different supervision strategies using a common action taxonomy, connecting performance changes to specific experimental behaviors.

\textbf{Decision-level supervision yields broad improvements in behavioral quality.} As shown in Figure~\ref{fig:agent-behavior}, Decision SFT achieves an overall quality pass rate of 96.11\%, exceeding Full SFT, step-level masking, and tool-failure masking at 87.25\%, 90.16\%, and 91.65\%, respectively, and outperforms Full SFT on 18 of the 20 action types. Since evaluation and checks account for a large share of records, we also recompute the aggregate after excluding this action type. The pass rates remain 95.63\% for Decision SFT and 87.40\% for Full SFT, indicating that quality improvements extend beyond evaluation activities.

\textbf{Quality improvements are particularly pronounced in key training choices.} Relative to Full SFT, Decision SFT raises the pass rates for learning objective, data source selection, and inference configuration from 79.27\%, 84.38\%, and 81.77\% to 95.05\%, 95.73\%, and 95.61\%, respectively. These choices determine how the target model learns, which data it uses, and how it performs inference, requiring judgments grounded in task requirements and experimental conditions. Their improved quality indicates that learning from trajectories benefits substantive choices about training approaches as well as operational execution. Gains vary across action types: pass rates for starting weights and storage management do not exceed those of Full SFT.

\textbf{The advantages over other selection strategies also appear in experimental behavior.} Decision SFT exceeds tool-failure masking by 10.49, 6.74, and 4.97 percentage points on inference configuration, sample generation, and data source selection, respectively, supporting the value of assessing behavior in its experimental context. Compared with step-level masking at a similar supervision volume, Decision SFT achieves higher pass rates on 17 action types, including gains of 9.49 and 7.40 percentage points on sample generation and learning objective. These differences support distinguishing step importance from behavioral quality: a step can be critical to task completion while its specific behavior remains unsuitable for supervision.

\section{Conclusion}
\label{sec:conclusion}

We introduced RSI-Prime, which combines complete model-training trajectories with fine-grained decision-quality annotations to help LLMs learn to train and improve other models. We apply decision-level loss masking to select high-quality behavior for supervision while preserving the full experimental context. PostTrainBench evaluation and behavioral analysis show that this approach improves agents' autonomous post-training performance and decision quality across multiple stages of experimentation, providing evidence for developing model-training capability through experimental experience.

\bibliography{custom}

\begin{thebibliography}{40}
\providecommand{\natexlab}[1]{#1}

\bibitem[{Chan et~al.(2024)Chan, Chowdhury, Jaffe, Aung, Sherburn, Mays,
  Starace, Liu, Maksin, Patwardhan, Weng, and Mądry}]{chan2024mlebench}
Jun~Shern Chan, Neil Chowdhury, Oliver Jaffe, James Aung, Dane Sherburn, Evan
  Mays, Giulio Starace, Kevin Liu, Leon Maksin, Tejal Patwardhan, Lilian Weng,
  and Aleksander Mądry. 2024.
\newblock \href {https://arxiv.org/abs/2410.07095} {{MLE-bench: Evaluating
  Machine Learning Agents on Machine Learning Engineering}}.
\newblock \emph{arXiv preprint arXiv:2410.07095}.

\bibitem[{Chen et~al.(2026)Chen, Yang, Yang, Sha, Li, Wang, Xian, Kong, Liu,
  and Bian}]{chen2026agent2rlbench}
Wanyi Chen, Xiao Yang, Xu~Yang, Tianming Sha, Qizheng Li, Zhuo Wang, Bowen
  Xian, Fang Kong, Weiqing Liu, and Jiang Bian. 2026.
\newblock \href {https://arxiv.org/abs/2604.10547} {{Agent}$^2$ {RL-Bench}: Can
  {LLM} agents engineer agentic {RL} post-training?}
\newblock \emph{Preprint}, arXiv:2604.10547.

\bibitem[{Chen et~al.(2025)Chen, Li, Huang, Du, Fang, and
  Zhou}]{chen-etal-2025-atlas}
Zhixun Chen, Ming Li, Yuxuan Huang, Yali Du, Meng Fang, and Tianyi Zhou. 2025.
\newblock \href {https://doi.org/10.18653/v1/2025.findings-acl.1299} {{ATLAS}:
  Agent tuning via learning critical steps}.
\newblock In \emph{Findings of the Association for Computational Linguistics:
  ACL 2025}, pages 25334--25349, Vienna, Austria. Association for Computational
  Linguistics.

\bibitem[{Du et~al.(2026{\natexlab{a}})Du, Yang, Zhou, Ge, Wang, Wang, Chen,
  Liu, Zhang, Gu, Zhang, Lei, and Chen}]{du2026rsimaster}
Yaxin Du, Xiyuan Yang, Zhifan Zhou, Yujie Ge, Cheng Wang, Jiajun Wang, Sijie
  Chen, Zehui Liu, Yuxin Zhang, Weicheng Gu, Julian Zhang, Zixing Lei, and
  Siheng Chen. 2026{\natexlab{a}}.
\newblock \href {https://arxiv.org/abs/2609.35561} {{RSI-Master}: Structuring
  experiments to guide autonomous model improvement}.
\newblock \emph{arXiv preprint arXiv:2609.35561}.

\bibitem[{Du et~al.(2026{\natexlab{b}})Du, Yang, Zhou, Liu, Lei, Chen, Liu, Wu,
  Cai, Liu, Zhu, Wang, Zhang, Qian, and Chen}]{du2026datamaster}
Yaxin Du, Xiyuan Yang, Zhifan Zhou, Wanxu Liu, Zixing Lei, Zimeng Chen, Fenyi
  Liu, Haotian Wu, Yuzhu Cai, Zexi Liu, Xinyu Zhu, WenHao Wang, Linfeng Zhang,
  Chen Qian, and Siheng Chen. 2026{\natexlab{b}}.
\newblock \href {https://arxiv.org/abs/2605.10906} {{DataMaster: Data-Centric
  Autonomous AI Research}}.
\newblock \emph{Preprint}, arXiv:2605.10906.

\bibitem[{Fang et~al.(2025)Fang, Han, Erickson, Zhang, Zhou, Dagar, Zhang,
  Turkmen, Hu, Rangwala, Wu, Wang, and Karypis}]{fang2025mlzero}
Haoyang Fang, Boran Han, Nick Erickson, Xiyuan Zhang, Su~Zhou, Anirudh Dagar,
  Jiani Zhang, Ali~Caner Turkmen, Cuixiong Hu, Huzefa Rangwala, Ying~Nian Wu,
  Bernie Wang, and George Karypis. 2025.
\newblock \href {https://arxiv.org/abs/2505.13941} {{MLZero: A Multi-Agent
  System for End-to-end Machine Learning Automation}}.
\newblock \emph{arXiv preprint arXiv:2505.13941}.

\bibitem[{Gao et~al.(2026{\natexlab{a}})Gao, Geng, Hua, Hu, Juan, Liu, Liu,
  Qiu, Qi, Wu, Wang, Xiao, Zhou, Zhang, Zhang, Xiang, Fang, Zhao, Liu, Ren,
  Qian, Wang, Hu, Wang, Wu, Ji, and Wang}]{gao2025selfevolving}
Huan-ang Gao, Jiayi Geng, Wenyue Hua, Mengkang Hu, Xinzhe Juan, Hongzhang Liu,
  Shilong Liu, Jiahao Qiu, Xuan Qi, Yiran Wu, Hongru Wang, Han Xiao, Yuhang
  Zhou, Shaokun Zhang, Jiayi Zhang, Jinyu Xiang, Yixiong Fang, Qiwen Zhao,
  Dongrui Liu, and 8 others. 2026{\natexlab{a}}.
\newblock \href {https://arxiv.org/abs/2507.21046} {{A Survey of Self-Evolving
  Agents: What, When, How, and Where to Evolve on the Path to Artificial Super
  Intelligence}}.
\newblock \emph{Preprint}, arXiv:2507.21046.

\bibitem[{Gao et~al.(2026{\natexlab{b}})Gao, Yu, Yin, Shi, Cai, He, and
  Lyu}]{gao2026rsidojo}
Shuzheng Gao, Zhaojian Yu, Penghao Yin, Yuling Shi, Kai Cai, Shilin He, and
  Michael~R. Lyu. 2026{\natexlab{b}}.
\newblock \href {https://arxiv.org/abs/2611.03698} {Training {LLM} agents to
  automate {LLM} post-training with {RSI-Dojo}}.
\newblock \emph{Preprint}, arXiv:2611.03698.

\bibitem[{Hu et~al.(2025)Hu, Lu, and Clune}]{hu2024adas}
Shengran Hu, Cong Lu, and Jeff Clune. 2025.
\newblock \href {https://arxiv.org/abs/2408.08435} {{Automated Design of
  Agentic Systems}}.
\newblock \emph{Preprint}, arXiv:2408.08435.

\bibitem[{Huang et~al.(2023)Huang, Vora, Liang, and
  Leskovec}]{huang2023mlagentbench}
Qian Huang, Jian Vora, Percy Liang, and Jure Leskovec. 2023.
\newblock \href {https://arxiv.org/abs/2310.03302} {{MLAgentBench: Evaluating
  Language Agents on Machine Learning Experimentation}}.
\newblock \emph{arXiv preprint arXiv:2310.03302}.

\bibitem[{Jiang et~al.(2025)Jiang, Schmidt, Srikanth, Xu, Kaplan, Jacenko, and
  Wu}]{jiang2025aide}
Zhengyao Jiang, Dominik Schmidt, Dhruv Srikanth, Dixing Xu, Ian Kaplan, Deniss
  Jacenko, and Yuxiang Wu. 2025.
\newblock \href {https://arxiv.org/abs/2502.13138} {{AIDE: AI-Driven
  Exploration in the Space of Code}}.
\newblock \emph{arXiv preprint arXiv:2502.13138}.

\bibitem[{Kang et~al.(2026)Kang, Li, Nguyen, Dabas, Ma, Sala, Song, and
  Jia}]{kang2026curationbench}
Feiyang Kang, Hanze Li, Adam Nguyen, Mahavir Dabas, Jiaqi~W. Ma, Frederic Sala,
  Dawn Song, and Ruoxi Jia. 2026.
\newblock \href {https://arxiv.org/abs/2606.04261} {{Can Generalist Agents
  Automate Data Curation?}}
\newblock \emph{Preprint}, arXiv:2606.04261.

\bibitem[{Liang et~al.(2023)Liang, Bommasani, Lee, Tsipras, Soylu, Yasunaga,
  Zhang, Narayanan, Wu, Kumar, Newman, Yuan, Yan, Zhang, Cosgrove, Manning, Re,
  Acosta-Navas, Hudson, Zelikman, Durmus, Ladhak, Rong, Ren, Yao, Wang,
  Santhanam, Orr, Zheng, Yuksekgonul, Suzgun, Kim, Guha, Chatterji, Khattab,
  Henderson, Huang, Chi, Xie, Santurkar, Ganguli, Hashimoto, Icard, Zhang,
  Chaudhary, Wang, Li, Mai, Zhang, and Koreeda}]{liang2023helm}
Percy Liang, Rishi Bommasani, Tony Lee, Dimitris Tsipras, Dilara Soylu,
  Michihiro Yasunaga, Yian Zhang, Deepak Narayanan, Yuhuai Wu, Ananya Kumar,
  Benjamin Newman, Binhang Yuan, Bobby Yan, Ce~Zhang, Christian Cosgrove,
  Christopher~D. Manning, Christopher Re, Diana Acosta-Navas, Drew~A. Hudson,
  and 31 others. 2023.
\newblock \href {https://openreview.net/forum?id=iO4LZibEqW} {Holistic
  evaluation of language models}.
\newblock \emph{Transactions on Machine Learning Research}.

\bibitem[{Lim et~al.(2026)Lim, Huang, Peng, Lu, Cong, Zhang, Sun, and
  Lin}]{lim2026missing}
Joy Jia~Yin Lim, Xin Huang, Hao Peng, Yaxi Lu, Xin Cong, Zhong Zhang, Maosong
  Sun, and Yankai Lin. 2026.
\newblock \href {https://arxiv.org/abs/2608.19072} {{What is Missing from AI
  Post-Training AI: An Empirical Analysis}}.
\newblock \emph{Preprint}, arXiv:2608.19072.

\bibitem[{Ma et~al.(2026)Ma, Wang, Xie, Chen, Du, Li, Sun, Liu, Chen, and
  Li}]{ma2026trex}
Zerun Ma, Guoqiang Wang, Xinchen Xie, Yicheng Chen, He~Du, Bowen Li, Yanan Sun,
  Wenran Liu, Kai Chen, and Yining Li. 2026.
\newblock \href {https://arxiv.org/abs/2604.14116} {{TREX: Automating LLM
  Fine-tuning via Agent-Driven Tree-based Exploration}}.
\newblock \emph{Preprint}, arXiv:2604.14116.

\bibitem[{Meng et~al.(2026)Meng, Du, Chen, Zhao, Lu, Hu, and
  Shieh}]{meng2026rsibenchdata}
Fanqing Meng, Lingxiao Du, Qiguang Chen, Ziqi Zhao, Haocheng Lu, Mengkang Hu,
  and Michael~Qizhe Shieh. 2026.
\newblock \href {https://arxiv.org/abs/2607.25886} {{RSIBench-Data}:
  Benchmarking data-centric research for recursive self-improvement}.
\newblock \emph{Preprint}, arXiv:2607.25886.

\bibitem[{{PostTrainBench
  Team}(2026{\natexlab{a}})}]{posttrainbench2026leaderboard}
{PostTrainBench Team}. 2026{\natexlab{a}}.
\newblock \href
  {https://github.com/aisa-group/posttrainbench-website/blob/main/scores-v1.2.json}
  {{PostTrainBench v1.2 Leaderboard Data}}.
\newblock Accessed November 5, 2026.

\bibitem[{{PostTrainBench Team}(2026{\natexlab{b}})}]{posttrainbench2026v12}
{PostTrainBench Team}. 2026{\natexlab{b}}.
\newblock \href {https://posttrainbench.com/blog/posttrainbench-1-2/}
  {{PostTrainBench v1.2: More Reliable Scores and Verdicts}}.

\bibitem[{Rank et~al.(2026)Rank, Bhatnagar, Prabhu, Eisenberg, Nguyen, Bethge,
  and Andriushchenko}]{rank2026posttrainbench}
Ben Rank, Hardik Bhatnagar, Ameya Prabhu, Shira Eisenberg, Karina Nguyen,
  Matthias Bethge, and Maksym Andriushchenko. 2026.
\newblock \href {https://arxiv.org/abs/2603.08640} {{PostTrainBench}: Can {LLM}
  agents automate {LLM} post-training?}
\newblock \emph{arXiv preprint arXiv:2603.08640}.

\bibitem[{Robeyns et~al.(2025)Robeyns, Szummer, and
  Aitchison}]{robeyns2025sica}
Maxime Robeyns, Martin Szummer, and Laurence Aitchison. 2025.
\newblock \href {https://arxiv.org/abs/2504.15228} {A self-improving coding
  agent}.
\newblock \emph{Preprint}, arXiv:2504.15228.

\bibitem[{Shinn et~al.(2023)Shinn, Cassano, Berman, Gopinath, Narasimhan, and
  Yao}]{shinn2023reflexion}
Noah Shinn, Federico Cassano, Edward Berman, Ashwin Gopinath, Karthik
  Narasimhan, and Shunyu Yao. 2023.
\newblock \href {https://arxiv.org/abs/2303.11366} {{Reflexion: Language Agents
  with Verbal Reinforcement Learning}}.
\newblock \emph{Preprint}, arXiv:2303.11366.

\bibitem[{Trirat et~al.(2024)Trirat, Jeong, and Hwang}]{trirat2024automlagent}
Patara Trirat, Wonyong Jeong, and Sung~Ju Hwang. 2024.
\newblock \href {https://arxiv.org/abs/2410.02958} {{AutoML-Agent: A
  Multi-Agent LLM Framework for Full-Pipeline AutoML}}.
\newblock \emph{arXiv preprint arXiv:2410.02958}.

\bibitem[{Wang et~al.(2023)Wang, Xie, Jiang, Mandlekar, Xiao, Zhu, Fan, and
  Anandkumar}]{wang2023voyager}
Guanzhi Wang, Yuqi Xie, Yunfan Jiang, Ajay Mandlekar, Chaowei Xiao, Yuke Zhu,
  Linxi Fan, and Anima Anandkumar. 2023.
\newblock \href {https://arxiv.org/abs/2305.16291} {{Voyager: An Open-Ended
  Embodied Agent with Large Language Models}}.
\newblock \emph{Preprint}, arXiv:2305.16291.

\bibitem[{Wang et~al.(2022)Wang, Mishra, Alipoormolabashi, Kordi, Mirzaei,
  Naik, Ashok, Dhanasekaran, Arunkumar, Stap, Pathak, Karamanolakis, Lai,
  Purohit, Mondal, Anderson, Kuznia, Doshi, Pal, Patel, Moradshahi, Parmar,
  Purohit, Varshney, Kaza, Verma, Puri, Karia, Doshi, Sampat, Mishra, Reddy~A,
  Patro, Dixit, and Shen}]{wang2022supernaturalinstructions}
Yizhong Wang, Swaroop Mishra, Pegah Alipoormolabashi, Yeganeh Kordi, Amirreza
  Mirzaei, Atharva Naik, Arjun Ashok, Arut~Selvan Dhanasekaran, Anjana
  Arunkumar, David Stap, Eshaan Pathak, Giannis Karamanolakis, Haizhi Lai,
  Ishan Purohit, Ishani Mondal, Jacob Anderson, Kirby Kuznia, Krima Doshi,
  Kuntal~Kumar Pal, and 16 others. 2022.
\newblock \href {https://doi.org/10.18653/v1/2022.emnlp-main.340}
  {Super-{N}atural{I}nstructions: Generalization via declarative instructions
  on 1600+ {NLP} tasks}.
\newblock In \emph{Proceedings of the 2022 Conference on Empirical Methods in
  Natural Language Processing}, pages 5085--5109. Association for Computational
  Linguistics.

\bibitem[{Yamada et~al.(2025)Yamada, Lange, Lu, Hu, Lu, Foerster, Clune, and
  Ha}]{yamada2025aiscientistv2}
Yutaro Yamada, Robert~Tjarko Lange, Cong Lu, Shengran Hu, Chris Lu, Jakob
  Foerster, Jeff Clune, and David Ha. 2025.
\newblock \href {https://arxiv.org/abs/2504.08066} {{The AI Scientist-v2:
  Workshop-Level Automated Scientific Discovery via Agentic Tree Search}}.
\newblock \emph{arXiv preprint arXiv:2504.08066}.

\bibitem[{Yao et~al.(2026)Yao, Wang, Lei, Luo, Zhang, Wu, Xu, Wang, Zhan, Guo,
  Zhang, Xu, Liu, and Xiong}]{yao2026improveanytask}
Xingbo Yao, Xiaoman Wang, Zhengwu Lei, Tinghui Luo, YiLin Zhang, Yuefeng Wu,
  Yijie Xu, Tianfu Wang, Qingyuan Zhan, Ye~Guo, Daoxin Zhang, Zhe Xu, Jian Liu,
  and Hui Xiong. 2026.
\newblock \href {https://arxiv.org/abs/2610.06347} {{ImproveAnyTask: An
  Autonomous Post-Training Harness for Iterative Model Self-Improvement}}.
\newblock \emph{Preprint}, arXiv:2610.06347.

\bibitem[{Yu et~al.(2026)Yu, Yin, Gao, He, Cai, and Zhang}]{yu2026autotrainess}
Zhaojian Yu, Penghao Yin, Shuzheng Gao, Shilin He, Kai Cai, and Xiao-Ping
  Zhang. 2026.
\newblock \href {https://arxiv.org/abs/2606.31551} {{AutoTrainess: Teaching
  Language Models to Improve Language Models Autonomously}}.
\newblock \emph{Preprint}, arXiv:2606.31551.

\bibitem[{Zhai et~al.(2025)Zhai, Tao, Chen, Zou, Chen, Fu, Mai, Yu, Deng, Cao,
  Liu, Ding, and Zhou}]{zhai2025agentevolver}
Yunpeng Zhai, Shuchang Tao, Cheng Chen, Anni Zou, Ziqian Chen, Qingxu Fu,
  Shinji Mai, Li~Yu, Jiaji Deng, Zouying Cao, Zhaoyang Liu, Bolin Ding, and
  Jingren Zhou. 2025.
\newblock \href {https://arxiv.org/abs/2511.10395} {{AgentEvolver: Towards
  Efficient Self-Evolving Agent System}}.
\newblock \emph{Preprint}, arXiv:2511.10395.

\bibitem[{Zhang et~al.(2026{\natexlab{a}})Zhang, Zhang, Li, Zhang, Chen, Zhang,
  Bai, and Hu}]{zhang2026selfharness}
Hangfan Zhang, Shao Zhang, Kangcong Li, Chen Zhang, Yang Chen, Yiqun Zhang, Lei
  Bai, and Shuyue Hu. 2026{\natexlab{a}}.
\newblock \href {https://arxiv.org/abs/2606.09498} {{Self-Harness: Harnesses
  That Improve Themselves}}.
\newblock \emph{Preprint}, arXiv:2606.09498.

\bibitem[{Zhang et~al.(2026{\natexlab{b}})Zhang, Hu, Lu, Lange, and
  Clune}]{zhang2025dgm}
Jenny Zhang, Shengran Hu, Cong Lu, Robert Lange, and Jeff Clune.
  2026{\natexlab{b}}.
\newblock \href {https://arxiv.org/abs/2505.22954} {{Darwin Godel Machine:
  Open-Ended Evolution of Self-Improving Agents}}.
\newblock \emph{Preprint}, arXiv:2505.22954.

\bibitem[{Zhang et~al.(2026{\natexlab{c}})Zhang, Zhao, Yang, Foerster, Clune,
  Jiang, Devlin, and Shavrina}]{zhang2026hyperagents}
Jenny Zhang, Bingchen Zhao, Wannan Yang, Jakob Foerster, Jeff Clune, Minqi
  Jiang, Sam Devlin, and Tatiana Shavrina. 2026{\natexlab{c}}.
\newblock \href {https://arxiv.org/abs/2603.19461} {{Hyperagents}}.
\newblock \emph{Preprint}, arXiv:2603.19461.

\bibitem[{Zhang et~al.(2026{\natexlab{d}})Zhang, Chen, Liu, Xue, Liao, Liu,
  Wang, Ning, Chen, Fu, Xie, Sun, Gou, Qi, Meng, Yang, Zhang, Li, Shah, Huynh,
  Li, Yang, Cao, Jang, Zhou, Zhu, Sun, Weston, Su, and
  Wu}]{zhang2025earlyexperience}
Kai Zhang, Xiangchao Chen, Bo~Liu, Tianci Xue, Zeyi Liao, Zhihan Liu, Xiyao
  Wang, Yuting Ning, Zhaorun Chen, Xiaohan Fu, Jian Xie, Yuxuan Sun, Boyu Gou,
  Qi~Qi, Zihang Meng, Jianwei Yang, Ning Zhang, Xian Li, Ashish Shah, and 11
  others. 2026{\natexlab{d}}.
\newblock \href {https://arxiv.org/abs/2510.08558} {{Agent Learning via Early
  Experience}}.
\newblock \emph{Preprint}, arXiv:2510.08558.

\bibitem[{Zhang et~al.(2026{\natexlab{e}})Zhang, Hu, Upasani, Ma, Hong,
  Kamanuru, Rainton, Wu, Ji, Li, Thakker, Zou, and Olukotun}]{zhang2025ace}
Qizheng Zhang, Changran Hu, Shubhangi Upasani, Boyuan Ma, Fenglu Hong,
  Vamsidhar Kamanuru, Jay Rainton, Chen Wu, Mengmeng Ji, Hanchen Li, Urmish
  Thakker, James Zou, and Kunle Olukotun. 2026{\natexlab{e}}.
\newblock \href {https://arxiv.org/abs/2510.04618} {{Agentic Context
  Engineering: Evolving Contexts for Self-Improving Language Models}}.
\newblock \emph{Preprint}, arXiv:2510.04618.

\bibitem[{Zhang et~al.(2024)Zhang, Tang, Wu, Wang, Shen, Hou, Tan, Li, Zhuang,
  and Lu}]{zhang2024agentpro}
Wenqi Zhang, Ke~Tang, Hai Wu, Mengna Wang, Yongliang Shen, Guiyang Hou, Zeqi
  Tan, Peng Li, Yueting Zhuang, and Weiming Lu. 2024.
\newblock \href {https://arxiv.org/abs/2402.17574} {{Agent-Pro}: Learning to
  evolve via policy-level reflection and optimization}.
\newblock \emph{Preprint}, arXiv:2402.17574.

\bibitem[{Zhao et~al.(2024)Zhao, Huang, Xu, Lin, Liu, and
  Huang}]{zhao2023expel}
Andrew Zhao, Daniel Huang, Quentin Xu, Matthieu Lin, Yong-Jin Liu, and Gao
  Huang. 2024.
\newblock \href {https://arxiv.org/abs/2308.10144} {{ExpeL: LLM Agents Are
  Experiential Learners}}.
\newblock \emph{Preprint}, arXiv:2308.10144.

\bibitem[{Zhao et~al.(2026{\natexlab{a}})Zhao, Ye, Ma, Liang, Feng, and
  Zhang}]{zhao2026andes}
Zhengyang Zhao, Shengjie Ye, Lu~Ma, Hao Liang, Hengyi Feng, and Wentao Zhang.
  2026{\natexlab{a}}.
\newblock \href {https://arxiv.org/abs/2606.01279} {{ANDES: Agent Native Data
  Evolving Synthesis Tool for Autonomous Instruction Alignment}}.
\newblock \emph{Preprint}, arXiv:2606.01279.

\bibitem[{Zhao et~al.(2026{\natexlab{b}})Zhao, Meng, Lu, Du, Chen, Hu, and
  Wu}]{zhao2026godelforest}
Ziqi Zhao, Fanqing Meng, Haocheng Lu, Lingxiao Du, Qiguang Chen, Mengkang Hu,
  and Xiao-Ming Wu. 2026{\natexlab{b}}.
\newblock \href {https://arxiv.org/abs/2609.36675} {G{\"o}del forest: Balancing
  search depth and breadth for data-centric recursive self-improvement}.
\newblock \emph{Preprint}, arXiv:2609.36675.

\bibitem[{Zhou et~al.(2026)Zhou, Guo, Liu, Yu, Gong, Zhao, Chen, Zhang, Chen,
  Li, Yang, Liu, Yu, Zhou, Wang, Sun, and Wang}]{zhou2026mementoskills}
Huichi Zhou, Siyuan Guo, Anjie Liu, Zhongwei Yu, Ziqin Gong, Bowen Zhao, Zhixun
  Chen, Menglong Zhang, Yihang Chen, Jinsong Li, Runyu Yang, Qiangbin Liu,
  Xinlei Yu, Jianmin Zhou, Na~Wang, Chunyang Sun, and Jun Wang. 2026.
\newblock \href {https://arxiv.org/abs/2603.18743} {{Memento-Skills}: Let
  agents design agents}.
\newblock \emph{Preprint}, arXiv:2603.18743.

\bibitem[{Zhou et~al.(2025)Zhou, Levine, Weston, Li, and
  Sukhbaatar}]{zhou2025selfchallenging}
Yifei Zhou, Sergey Levine, Jason Weston, Xian Li, and Sainbayar Sukhbaatar.
  2025.
\newblock \href {https://arxiv.org/abs/2506.01716} {Self-challenging language
  model agents}.
\newblock \emph{Preprint}, arXiv:2506.01716.

\bibitem[{Zweiger et~al.(2025)Zweiger, Pari, Guo, Akyürek, Kim, and
  Agrawal}]{zweiger2025seal}
Adam Zweiger, Jyothish Pari, Han Guo, Ekin Akyürek, Yoon Kim, and Pulkit
  Agrawal. 2025.
\newblock \href {https://arxiv.org/abs/2506.10943} {{Self-Adapting Language
  Models}}.
\newblock \emph{Preprint}, arXiv:2506.10943.

\end{thebibliography}

\clearpage
\appendix
\twocolumn[{\centering\Large\bfseries Appendix\par\vspace{1em}}]
\section{Additional Dataset Analysis}
\label{sec:dataset-details}

\subsection{Analysis Scope and Action Coverage}

The dataset comprises 350 benchmarks, 350 trajectories, and 385 training segments. The annotation analysis covers 90,738 decision records. Decision records capture identified choices and their cross-window continuations. Of these records, 90,650 map to at least one assistant message; 49,862 map to multiple messages. Mapped records reference 1.90 messages on average. Among 80,090 distinct referenced assistant messages, 45,661 are referenced by multiple records. Tool results are associated with the calling assistant message using tool-call identifiers. Table~\ref{tab:action-coverage} summarizes the distribution of decision records across action categories.

\begin{table}[htbp]
  \centering
  \small
  \setlength{\tabcolsep}{3pt}
  \begin{tabular}{@{}lrr@{}}
    \toprule
    \textbf{Action category} & \textbf{Records} & \textbf{Share (\%)} \\
    \midrule
    Evaluation and diagnosis & 34,872 & 38.43 \\
    Data preparation & 24,583 & 27.09 \\
    Runtime support & 16,194 & 17.85 \\
    Model training & 8,252 & 9.09 \\
    Model composition and delivery & 6,564 & 7.23 \\
    Other & 273 & 0.30 \\
    \bottomrule
  \end{tabular}
  \caption{Action distribution over decision records. Rounded shares need not sum to exactly 100\%.}
  \label{tab:action-coverage}
\end{table}

The quality analysis groups decision records into 83,856 high-quality and 6,882 problematic records, accounting for 92.42\% and 7.58\%, respectively. We count both categories within each trajectory to characterize the coexistence of different-quality decisions.

\subsection{Illustrative Problem Patterns}
\label{sec:dataset-patterns}

We inspect source trajectories with model assistance to identify recurring problems in data construction, evaluation startup, and environment repair.

\paragraph{Data construction does not implement selected requirements.}
Four trajectories illustrate mismatches between a chosen data plan and its implementation. In TOD-ProcBench, the constructed mixture differs from the selected 55:45 ratio. In ManyIH, the current generation job covers only one of the selected instruction groups. In XStory, the filtering rules exclude most of the intended training source. In FoFo, the script omits a planned correction-and-regeneration step.

\paragraph{Waiting without confirming evaluation startup.}
In CLINC150 and TAT-QA, the agent waits after receiving a background submission receipt, subsequently finds the expected evaluation log missing, and restarts evaluation. Both cases show that the agent waited without first confirming that evaluation had started.

\paragraph{Repairs preserve a known failure condition.}
In Schema-Guided Dialog, the agent continues downloading a large dependency over a known slow connection despite a locally available copy. In SummScreen, command variants retain the same interpreter and leave the CUDA dependency issue unresolved. Both illustrate environment-repair attempts that fail to change an identified obstacle.

\subsection{Supervision Accounting}
\label{sec:supervision-details}

Decision SFT supervises only messages linked to decisions that pass quality selection. Under the supervision selection rule, a decision passes when all three quality dimensions are labeled reasonable or uncertain. Problematic, not-applicable, or no-majority outcomes make a decision ineligible. If any linked decision fails, the entire assistant message is masked; unlinked messages are also masked. For the binary quality summary in Section~\ref{sec:dataset-analysis}, high-quality records require reasonable analysis and choice, with execution labeled reasonable or uncertain; all remaining records are grouped as problematic.

Decision-level masking retains 68,027 of 89,211 originally supervised assistant messages and 33,477,528 of 44,353,402 target tokens, yielding retention rates of 76.25\% and 75.48\%, respectively. Token counts use the fixed local renderer and GLM-5.3 tokenizer.

Masked content includes 12,094 unlinked messages containing 956,754 target tokens. Such content is not labeled problematic merely because it lacks a decision association. Other masking triggers include problematic labels, non-applicability, and unresolved majority votes; masked token counts therefore do not estimate the amount of erroneous content.

Action-specific retention uses the union of messages associated with each action, so categories can overlap and their token counts are not additive. Table~\ref{tab:action-supervision} reports retention within these linked messages.

\begin{table}[!htbp]
  \centering
  \small
  \setlength{\tabcolsep}{3pt}
  \begin{tabular}{@{}lrr@{}}
    \toprule
    \textbf{Action category} & \textbf{Messages} & \textbf{Tokens} \\
    \midrule
    Data preparation & 80.21 & 69.38 \\
    Model training & 79.56 & 68.86 \\
    Evaluation and diagnosis & 87.77 & 76.18 \\
    Model composition and delivery & 84.27 & 72.31 \\
    Runtime support & 85.01 & 73.61 \\
    Other & 68.66 & 58.41 \\
    \bottomrule
  \end{tabular}
  \caption{Supervision retention (\%) within each action category, calculated over linked messages.}
  \label{tab:action-supervision}
\end{table}

\subsection{Human Audit of Decision Annotations}
\label{sec:human-validation}

We conduct a human audit to assess the reliability of decision identification, action classification, and quality assessment. We randomly sample 400 decision records without replacement from the annotated corpus. Each record is presented with its original source messages and relevant context, the identified decision and associated evidence, the assigned action labels, and the quality judgments with their supporting rationales.

Each record is independently reviewed by two annotators. For decision identification, reviewers assess whether the decision is supported by the source and whether its organization and evidence attribution are appropriate. For action classification, they assess whether the assigned labels characterize the core choice. For quality assessment, they examine whether the labels and rationales are justified by the available evidence, respecting the information available at the time of the decision. Reviewers provide a binary acceptance judgment for each audit dimension. Disagreements are adjudicated by a third reviewer, while initial agreements remain unchanged.

We report acceptance rates after adjudication and inter-annotator agreement before adjudication. Raw agreement and Cohen's $\kappa$ are computed separately for each dimension from the paired initial judgments over the 400 records (Table~\ref{tab:human-audit}).

\begin{table}[!htbp]
  \centering
  \small
  \setlength{\tabcolsep}{3pt}
  \begin{tabular}{@{}lrrr@{}}
    \toprule
    \textbf{Audit dimension} & \shortstack{\textbf{Acceptance}\\\textbf{(\%)}} & \shortstack{\textbf{Agreement}\\\textbf{(\%)}} & $\boldsymbol{\kappa}$ \\
    \midrule
    Decision identification & 98.75 & 99.50 & 0.831 \\
    Action classification & 99.50 & 99.75 & 0.799 \\
    Quality assessment & 97.50 & 99.25 & 0.866 \\
    \bottomrule
  \end{tabular}
  \caption{Human audit of 400 randomly sampled decision records. Acceptance rates use adjudicated judgments; raw agreement and Cohen's $\kappa$ use the initial independent judgments.}
  \label{tab:human-audit}
\end{table}

The audit yields high acceptance rates across all three dimensions. After adjudication, reviewers accept 395 decision identifications, 398 action classifications, and 390 quality assessments. Initial disagreements occur in two, one, and three records, respectively, and all are adjudicated as acceptable. Action classification has the highest raw agreement but a lower $\kappa$ because the acceptance judgments are more heavily concentrated in one category.

These results support the reliability of the sampled annotations. Since the audit samples existing decision records, it does not estimate the recall of decision identification or the frequency of omitted decisions.

\section{Reference Agent Evaluation}
\label{sec:baseline-runs}

We evaluate the original GLM-5.3 agent using DSH under the evaluation settings in Section~\ref{sec:experimental-setup}. Each experiment trains a target model initialized from Qwen3-4B-Base with a cumulative 10-hour agent budget. Table~\ref{tab:baseline-runs} reports three valid independent runs per benchmark and their arithmetic mean. The means form the GLM-5.3 reference row in Table~\ref{tab:experimental-results}; applying the official full-precision benchmark weights to the unrounded means gives an Overall score of 36.56. Run indices are local to each benchmark and do not identify matched runs across the six tasks. This three-run averaging applies only to the original GLM-5.3 agent.

\begin{table}[htbp]
  \centering
  \small
  \setlength{\tabcolsep}{3pt}
  \renewcommand{\arraystretch}{1.08}
  \begin{tabular}{@{}lrrrr@{}}
    \toprule
    \textbf{Benchmark} & \textbf{Run 1} & \textbf{Run 2} & \textbf{Run 3} & \textbf{Mean} \\
    \midrule
    AIME 2025 & 13.33 & 13.33 & 16.67 & 14.44 \\
    Arena Hard Writing & 25.79 & 37.03 & 52.57 & 38.46 \\
    GPQA Main & 27.23 & 28.79 & 30.80 & 28.94 \\
    GSM8K & 77.86 & 84.15 & 88.93 & 83.65 \\
    HealthBench & 20.72 & 23.64 & 26.39 & 23.58 \\
    HumanEval & 78.66 & 79.27 & 79.88 & 79.27 \\
    \bottomrule
  \end{tabular}
  \caption{Our evaluation of the original GLM-5.3 agent: individual task scores and means over three valid independent runs per benchmark.}
  \label{tab:baseline-runs}
\end{table}

\end{document}